# Understanding and Improving UMAP with Geometric and Topological Priors: The JORC-UMAP Algorithm

Xiaobin Li and Run Zhang

School of Mathematics, Southwest Jiaotong University,

West zone, High-tech district, Chengdu, Sichuan 611756, China

January 23, 2026

## Abstract

Nonlinear dimensionality reduction techniques, particularly UMAP, are widely used for visualizing high-dimensional data. However, UMAP's local Euclidean distance assumption often fails to capture intrinsic manifold geometry, leading to topological tearing and structural collapse. We identify UMAP's sensitivity to the k-nearest neighbor graph as a key cause. To address this, we introduce Ollivier-Ricci curvature as a geometric prior, reinforcing edges at geometric bottlenecks and reducing redundant links. Since curvature estimation is noise-sensitive, we also incorporate a topological prior using Jaccard similarity to ensure neighborhood consistency. The resulting method, JORC-UMAP, better distinguishes true manifold structure from spurious connections. Experiments on synthetic and real-world datasets show that JORC-UMAP reduces tearing and collapse more effectively than standard UMAP and other DR methods, as measured by SVM accuracy and triplet preservation scores, while maintaining computational efficiency. This work offers a geometry-aware enhancement to UMAP for more faithful data visualization.

## 1 Introduction

Dimensionality reduction (DR) has become an indispensable tool for exploring high- dimensional data across numerous scientific fields. Its fundamental aim is to embed such data into a low-dimensional representation while preserving its intrinsic structure as faithfully as possible [7]. This method has found widespread adoption in diverse domains, including computer vision [29], single-cell biology [10], and natural language processing [1]. Motivated by the demands of practical applications, a broad class of nonlinear dimensionality reduction methods has been proposed to more effectively characterize both the local and global structures of complex data. In recent years, UMAP (Uniform Manifold Approximation and Projection) [15] has demonstrated clear advantages over t-SNE [14] in terms of structural preservation and computational efficiency across a wide range of scenarios, and has consequently emerged as a prominent approach within the field of nonlinear dimensionality reduction.

Despite its advantages, UMAP's performance is contingent on the quality of its initial k-nearest neighbor graph, which is constructed based on local Euclidean distances. This approach implicitly assumes local uniformity and flatness of the data manifold, an assumption that often breaks down in the presence of significant curvature or complex topology. Specifically, UMAP primarily relies on local Euclidean distances to construct the k-nearest neighbor graph and implicitly assumes that the manifold can be approximated as a flat Euclidean space within sufficiently small neighborhoods. Consequently, the characterization of local neighborhood relationships is essentially conducted under a Euclidean metric framework, which amounts to treating local structural information as if it were embedded in a Euclidean space. However, when the manifold exhibits nonzero curvature or contains thin connective structures, this assumption may fail to adequately reflect the influence of curvature on local distance estimation, potentially leading to distortions in the resulting low-dimensional embedding. Such distortions mainly manifest in the following two forms:

1. **Topological tearing**: the "skeleton"of the manifold or thin connective regions may be disconnected due to slightly overestimated local distances, causing originally connected clusters to separate in the low-dimensional space.

2. **Structural collapse**: in regions with high curvature, insufficient characterization of local geometric information may result in unreasonable contraction or distortion in the embedding.

While manifesting differently, both topological tearing and structural collapse share a common origin: the failure of the Euclidean-based k-nearest neighbor graph to accurately capture the intrinsic geometric and connective properties of the underlying data manifold, particularly in regions of high curvature. In Section 4, we will conduct experimental analyses on representative manifold datasets to empirically illustrate the manifestation of these topological deformation risks in standard UMAP.

To alleviate the aforementioned issues, this work considers and exploits the discrete curvature information of the data, specifically using the Ollivier-Ricci curvature (ORC) [19]. This notion of Ollivier—Ricci curvature is grounded in optimal transport theory and quantifies local bending in discrete metric spaces by comparing the Wasserstein distance between neighborhood probability distributions relative to the geodesic distance between nodes. In general, edges with negative curvature are often associated with bottleneck structures or hub-like connections on the manifold, whereas positive curvature typically arises within the interior of high-density clusters.Nevertheless, the direct application of ORC still faces a critical robustness challenge, namely the confusion between noise and genuine structural features [17]. Since ORC evaluates transport costs between neighborhoods via the Wasserstein distance, in high-dimensional settings noisy edges may inadvertently connect two originally disjoint local neighborhoods, thereby inducing a large transport cost. As a result, such edges may numerically exhibit negative curvature characteristics similar to those of true manifold skeletons.

The application of ORC introduces a new challenge: its sensitivity to noise. In high-dimensional spaces, spurious connections can exhibit transport properties similar to genuine manifold bottlenecks, leading to false negative curvature readings. To robustly distinguish true structural edges from noise-induced shortcuts, we introduce a topological prior via the Jaccard similarity coefficient. We observe that, in practical datasets, authentic manifold skeletons typically exhibit a higher degree of local neighborhood overlap, whereas connections arising from noise tend to be more random and are often accompanied by neighborhood sparsity. Based on this observation, the Jaccard similarity coefficient [9] is introduced to quantify the overlap between the neighbor sets of nodes at the two ends of an edge, and is employed to filter potential noise-induced connections, thereby enhancing the robustness of discrete curvature estimation under noisy conditions. Building upon this mechanism, we propose a dimensionality reduction algorithm within the UMAP framework that integrates Ollivier-Ricci curvature with the Jaccard similarity coefficient, termed JORC-UMAP (Jaccard Ollivier-Ricci Curvature UMAP).

The main contributions of this work can be summarized as follows:

1. **Geometric Prior Integration** : We incorporate Ollivier-Ricci curvature into UMAP's graph construction phase, enabling curvature-aware reweighting of edges to enhance structural preservation in high-curvature regions and bottlenecks.

2. **Topological Noise Filtering**: We propose a Jaccard similarity-based neighborhood overlap validation mechanism to filter spurious connections introduced by noise, preventing them from being erroneously identified by discrete curvature as structurally meaningful negative-curvature edges. This mechanism enables a clearer distinction between genuine manifold skeleton connections and noise-induced bridges.

3. **A Modular Framework**: The proposed JORC-UMAP operates at the k-nearest neighbor graph construction stage for high-dimensional data and constitutes a modification of UMAP , s graph structure. This idea is independent of the objective functions or optimization strategies adopted by methods such as TriMAP and PaCMAP, and can therefore be extended to those frameworks.

# 2 Related Work

This section reviews related work along three lines: the development of the UMAP algorithm, the theory and graph-learning applications of Ollivier-Ricci curvature, and the role of the Jaccard similarity coefficient in dimensionality reduction.

## 2.1 The Development of the UMAP Algorithm

The development of nonlinear dimensionality reduction (DR) has long been characterized by a fundamental trade-off: preserving local neighborhood structures versus maintaining global structural consistency. Early methods like t-SNE [14] made significant strides in visualization by minimizing the Kullback-Leibler divergence between high- and low-dimensional probability distributions. However, t-SNE's strong emphasis on local neighborhoods often comes at the cost of global geometric fidelity and computational scalability, limiting its effectiveness on large-scale datasets.

A major advance came with UMAP [15], which leverages principles from Riemannian geometry and algebraic topology. By constructing a fuzzy simplicial complex and optimizing a cross-entropy loss, UMAP achieves superior computational efficiency and scalability while effectively preserving local structure. Consequently, it has become an important method in nonlinear

DR. Nonetheless, a critical limitation remains: UMAP's capture of global structure is highly sensitive to the quality of the initial k-nearest neighbor graph.

To address this issue, TriMap [4] introduces triplet constraints to preserve global relative distance relationships, while PaCMAP [26] seeks a more favorable balance between local and global structures by dynamically adjusting the weights of near, mid-range, and far point pairs. Although these methods preserve more global structure to some extent, they remain fundamentally grounded in the statistical properties of Euclidean distances and lack an explicit representation of the intrinsic geometric curvature of the data.

### 2.2 Ollivier-Ricci Curvature and Its Applications in Graph Learning

In classical Riemannian geometry, Ricci curvature characterizes the deviation between the volume of geodesic balls on a manifold and that of corresponding balls in Euclidean space [18]. However, Ricci curvature is defined on smooth manifolds and relies on second-order differential tensors, which prevents its direct application to discrete data point clouds or graph structures. Consequently, it is necessary to introduce a notion of discrete curvature that does not depend on differential structures and is defined solely in terms of metric space properties. To this end, Yann Ollivier proposed a discrete curvature concept applicable to general metric spaces, including graphs, known as Ollivier-Ricci curvature (ORC) [19].

Because ORC exhibits strong discriminative capability for identifying negative-curvature edges, it has been successfully applied to tasks such as graph neural networks (GNNs) [16], community detection in complex networks [23], and biological system analysis [24]. In contrast, the systematic incorporation of ORC into the neighborhood graph construction stage of manifold learning and dimensionality reduction algorithms remains relatively limited. In this work, we investigate the integration of ORC into the weight computation process of UMAP in order to enhance its ability to preserve structural characteristics of the data. Unlike the ORC-MANL method [20], which removes redundant edges from the k-nearest neighbor graph based on ORC, we argue that negative-curvature edges often correspond to key structural pathways on the manifold, and that indiscriminate removal may disrupt global connectivity. Accordingly, we propose to selectively reinforce negative-curvature edges that carry structural significance.

### 2.3 The Jaccard Similarity Coefficient in Dimensionality Reduction Algorithms

The Jaccard index (Jaccard similarity coefficient) was originally introduced by Paul Jaccard as a measure of similarity between two sets [9]. In graph theory and complex network analysis, the Jaccard coefficient has been widely adopted as a similarity metric for characterizing neighborhood overlap, and has been extensively applied to tasks such as link prediction [13], community detection [30], and contrastive graph clustering [2].

In UMAP and other dimensionality reduction algorithms based on k-nearest neighbor graphs, erroneous connections introduced during the graph construction stage are often a major source of structural distortion in the resulting low-dimensional embeddings. Although ORC is capable of identifying geometric features of the manifold, such as bottlenecks and cluster centers, it is relatively sensitive to sparse noise and may misclassify noise-induced sparse connections as structurally meaningful edges with negative curvature. Consequently, incorporating the Jaccard index as a constraint independent of geometric curvature helps eliminate spurious negative-curvature connections, thereby improving the reliability of the dimensionality reduction results.

By combining the geometric sensitivity of ORC with the neighborhood overlap discrimination provided by the Jaccard index, JORC-UMAP aims to overcome the limitations of relying on a single criterion and to provide a dimensionality reduction approach that more faithfully reflects the intrinsic structure of the data.

# 3 Principles of the JORC-UMAP Algorithm

## 3.1 Overview of UMAP Algorithm

UMAP assumes that the data are distributed on a low-dimensional Riemannian manifold ($M^d$, g) embedded in a high-dimensional Euclidean space $R^m$, where $d \ll m$. It further assumes that, within sufficiently small local neighborhoods, the data are uniformly distributed with respect to an approximately constant metric tensor. Under these assumptions, UMAP constructs a neighborhood graph from the high-dimensional data to characterize the local geometric structure, and subsequently derives a low-dimensional embedding based on this representation.

The algorithm can be broadly divided into two stages. The first stage is graph construction: for each data point, k-nearest neighbor relationships are established based on a local distance metric (typically the Euclidean distance), and the distances are mapped to probabilistic edge weights via adaptive scaling parameters. This yields a symmetric weighted neighborhood graph, which can be viewed as an approximate representation of the local fuzzy topological structure of data.The second stage is embedding optimization: in the low-dimensional space, the cross-entropy between the weighted graphs in the high- and low-dimensional spaces is minimized, so that points with high similarity in the high-dimensional space are encouraged to attract each other in the low-dimensional embedding, while non-neighboring point pairs exert repulsive forces, ultimately forming the final embedding.

From an optimization perspective, this process can be interpreted as a force-directed graph layout method optimized via stochastic gradient descent, where neighboring edges correspond to attractive forces and non-adjacent point pairs give rise to repulsive forces. The expressive power and stability of UMAP depend to a large extent on how well the initial neighborhood graph captures the local geometric structure of the data, which in turn makes the graph construction stage a critical component for methodological improvement.

## 3.2 Ollivier Ricci Curvature Calculation Based on Optimal Transport Theory

When the data exhibit pronounced variations in curvature, the neighborhood graph constructed by UMAP based on local Euclidean distances may deviate from the true structure of the

underlying data manifold, thereby giving rise to the following two types of risks:

1. **Topological tearing**: the "skeleton" of the manifold or thin connective regions may be disconnected due to slightly overestimated local distances, causing originally connected clusters to separate in the low-dimensional space.

2. **Structural collapse**: in regions with high curvature, insufficient characterization of local geometric information may result in unreasonable contraction or distortion in the embedding.

To mitigate the aforementioned issues, we incorporates discrete curvature information of the data during the neighborhood graph construction stage.

Within the framework of optimal transport theory, Ollivier—Ricci curvature (Ollivier—Ricci Curvature, ORC) provides a discrete notion of curvature for general metric spaces. This curvature quantifies the local contraction or expansion of geometric structure by comparing the Wasserstein distance between neighborhood probability distributions of adjacent nodes, thereby enabling the incorporation of curvature information into k-nearest neighbor graphs. The computation of Ollivier–Ricci curvature can be summarized in the following three steps.

1. **Definition of local probability measures**: On the weighted k-nearest neighbor graph $G = (V, E)$, a local probability measure $m_{x_i}$ supported on the neighborhood of each node $x_i$ is defined. To remain consistent with the graph construction procedure in UMAP, the edge weights $w_{ij}$ are directly normalized as follows:

$$m_{x_i}(x_k) = \begin{cases} \alpha & \text{if } x_k = x_i \\ (1-\alpha)\frac{w_{ik}}{\sum_{v \in N(x_i)} w_{iv}} & \text{if } x_k \in N(x_i) \\ 0 & \text{otherwise} \end{cases}$$

where $N(x_i)$ denotes the k-nearest neighbor set of $x_i$, and $\alpha \in [0, 1]$ is a laziness parameter that controls the probability of the random walk remaining at the current node.

2. **Computation of the Wasserstein distance**: For any edge $(x_i, x_j)$, the Wasserstein distance between the corresponding measures $m_{x_i}$ and $m_{x_j}$ is computed as:

$$W_1(m_{x_i}, m_{x_j}) = \inf_{\gamma \in \Pi(m_{x_i}, m_{x_j})} \sum_{u,v \in V} d(u,v)\gamma(u,v)$$

where $d(u, v)$ denotes the shortest-path distance on the graph, and $\Pi(m_{x_i}, m_{x_j})$ represents the set of all joint probability distributions $\gamma$ with marginals $m_{x_i}$ and $m_{x_j}$. The computation of the Wasserstein distance constitutes a classical linear programming problem, whose exact solution incurs a time complexity as high as $O(k^3 \log k)$. To substantially reduce the computational cost, this work adopts the Sinkhorn—Knopp algorithm to obtain an approximate solution [3].

3. **Computation of Ollivier-Ricci curvature** Based on the Wasserstein distance obtained above, the Ollivier—Ricci curvature $\kappa(x_i, x_j)$ on an edge $(x_i, x_j)$ is defined as:

$$\kappa(x_i, x_j) = 1 - \frac{W_1(m_{x_i}, m_{x_j})}{d(x_i, x_j)}$$

The geometric and topological interpretations of this curvature value are as follows:

(a) **Positive curvature** ($\kappa > 0$): the transport distance between neighborhoods is smaller than the distance between the central nodes, indicating a locally convergent structure. This typically corresponds to the interior of high-density clusters, where connections exhibit high local redundancy.

(b) **Zero curvature** ($\kappa \approx 0$): the neighborhood structure is approximately Euclidean and locally flat, reflecting a relatively uniform data distribution.

(c) **Negative curvature** ($\kappa < 0$): the transport distance between neighborhoods exceeds the distance between the central nodes, indicating a locally divergent structure. This often corresponds to bottlenecks, skeletons, or inter-cluster connections on the manifold.

The curvature matrix $\kappa_{ij}$ obtained from the above procedure is incorporated into the graph construction stage of UMAP to adjust edge weights, thereby refining similarity modeling and enhancing the connectivity of key manifold skeleton structures.

## 3.3 Construction of the JORC-UMAP Model

Ollivier–Ricci curvature (ORC) is primarily used to characterize key structural connections within the data manifold, which typically correspond to negatively curved bottleneck regions or inter-cluster hubs. In the standard UMAP neighborhood graph construction process, such connections are often weakened or even lost due to biases in local distance estimation. However, relying solely on ORC for discrimination still suffers from limited robustness: noise points or outliers may induce negative-curvature responses similar to those of genuine structural connections, thereby leading to misclassification.

Specifically, genuine structural connections typically appear as two high-density regions linked by a narrow, low-density manifold channel. In such situations, the transport distance $W_1$ between the neighborhood distributions of adjacent nodes is substantially larger than the geodesic distance d between the nodes, resulting in a negative curvature value $\kappa < 0$ for the corresponding edge. These negative-curvature edges correspond to structural connections that should be preferentially preserved or reinforced.

In contrast, noise-induced bridges are often formed when a small number of anomalous points in the high-dimensional space accidentally connect otherwise unrelated clusters. Because the neighborhoods of these points are sparse and exhibit large distributional discrepancies, the associated $W_1$ may also be large, thereby producing negative curvature responses that are difficult to distinguish from those of genuine manifold skeletons. Consequently, indiscriminately enhancing all negative-curvature edges would inevitably amplify noise-induced connections, leading to erroneous aggregation of originally separate clusters in the low-dimensional embedding. To distinguish genuine structural connections from such spurious noise bridges, this work introduces the Jaccard similarity coefficient as a noise validation criterion in addition to the curvature constraint.

The Jaccard similarity coefficient J(x, y) is used to quantify the degree of overlap between the k-nearest neighbor sets N(x) and N(y) of nodes x and y.

$$J(x,y) = \frac{|N(x) \cap N(y)|}{|N(x) \cup N(y)|}$$

In JORC-UMAP, the Jaccard similarity coefficient is jointly employed with Ollivier-Ricci curvature to regulate the edge weights of the high-dimensional k-nearest neighbor graph. Specifically, edges are categorized into three types according to their curvature values and neighborhood overlap characteristics:

1. **Genuine skeleton edges**: edges with negative curvature and high neighborhood overlap, indicating true structural connections on the manifold. These edges are strengthened to preserve global connectivity.

2. **Noise connections**: edges with negative curvature but low neighborhood overlap, typically induced by random noise or outliers. Such edges are strongly suppressed to prevent spurious connections.

3. **Intra-cluster edges**: edges corresponding to locally convergent or approximately flat regions, which are retained with their original weights or subjected to mild suppression.

Accordingly, the updated weight of an edge (i, j) is defined as:

$$w'_{ij} = \begin{cases} w_{ij} + (1 - w_{ij}) \cdot \tanh(S \cdot |\kappa_{ij}|) & \text{if } \kappa < 0 \text{ and } J > \delta \text{ (genuine skeleton edges)} \\ w_{ij} \cdot (1 - \tanh(S \cdot \kappa_{ij}) \cdot \beta) & \text{if } \kappa \geq 0 \text{ (intra-cluster)} \\ w_{ij} \cdot \epsilon & \text{if } \kappa < 0 \text{ and } J \leq \delta \text{ (Noise connections)} \end{cases}$$

where δ denotes the Jaccard threshold used to determine whether neighborhood overlap is sufficient, and serves as a key hyperparameter for distinguishing structural skeleton edges from noise-induced bridge edges; $\epsilon$ is a small constant(e.g., $10^{-5}$), employed to suppress the weights of noise-induced connections to near zero; S is an intensity factor that controls the nonlinear response of curvature in the weight modulation, which can be adaptively set according to the statistical distribution of $|\kappa_{ij}|$ and is typically fixed to 2.0 when specified manually; the tanh(·) function provides a smooth and bounded mapping to avoid abrupt changes in edge weights; and $\beta \in [0, 1]$ is a suppression coefficient for intra-cluster edges, used to moderately weaken redundant connections in regions of positive curvature, with a default value of 0.9.

The Jaccard similarity coefficient provides an effective constraint for ORC, enabling genuine structural connections to be distinguished from noise-induced spurious links. The above procedure constitutes the core mechanism of JORC-UMAP, and its overall workflow, as summarized in Algorithm 1, includes neighborhood graph construction, Ollivier-Ricci curvature computation, Jaccard similarity evaluation, and subsequent weight updates based on the joint criteria.

**Algorithm 1** Jaccard Ollivier-Ricci Curvature UMAP (JORC-UMAP)

**Require:** Data $X \in \mathbb{R}^{N \times D}$, Neighbors $k$, Noise Threshold $\delta$, Target Tanh $T$
**Ensure:** Low-dimensional embedding $Y \in \mathbb{R}^{N \times d}$

1: **Initialize:** Construct fuzzy simplicial set graph $G = (V, E)$ with weights $W$ via UMAP.
2: $Y \leftarrow$ Initialization($X$)  ▷ PCA
   // *Phase 1: Curvature Quantification*
3: **for** each edge $(i, j) \in E$ **do**
4:     Compute local probability measures $m_{x_i}, m_{x_j}$ from $W$.
5:     $W_1(i, j) \leftarrow$ Sinkhorn($m_{x_i}, m_{x_j}$)  ▷ Optimal Transport Distance
6:     $\kappa_{ij} \leftarrow 1 - W_1(i, j)/\|x_i - x_j\|_2$  ▷ Ollivier-Ricci Curvature
7: **end for**
   // *Phase 2: Dynamic Strength Estimation*
8: **if** Dynamic Strength is True **then**
9:     $K_{neg} \leftarrow \{\kappa_{ij} \mid \kappa_{ij} < 0\}$
10:    $\kappa_{typ} \leftarrow$ Percentile($|K_{neg}|$, 75)  ▷ Robust statistic via 75th percentile
11:    $S \leftarrow \text{arctanh}(T)/(\kappa_{typ} + \epsilon)$  ▷ Targeting $\tanh(S \cdot \kappa_{typ}) \approx T$
12:    $S \leftarrow$ Clip($S$, 0.5, 10.0)
13:    Adjust $S$ based on ratio $|K_{neg}|/|E|$  ▷ Scaling factor 1.5 or 0.8
14: **else**
15:    $S \leftarrow S_{base}$
16: **end if**
   // *Phase 3: Topology Rectification (Reweighting)*
17: **for** each edge $(i, j) \in E$ **do**
18:    $J_{ij} \leftarrow |N(i) \cap N(j)|/|N(i) \cup N(j)|$  ▷ Jaccard Similarity
19:    **if** $\kappa_{ij} < 0$ **then**  ▷ Possible manifold tear/bridge
20:        **if** $J_{ij} < \delta$ **then**
21:            $w'_{ij} \leftarrow w_{ij} \cdot 10^{-5}$  ▷ **Cut Noise**: False bridge detected
22:        **else**
23:            boost $\leftarrow \tanh(S \cdot |\kappa_{ij}|)$
24:            $w'_{ij} \leftarrow w_{ij} + (1 - w_{ij}) \cdot$ boost  ▷ **Lock Skeleton**: Enhance connectivity
25:        **end if**
26:    **else**  ▷ Cluster interior
27:        suppress $\leftarrow \tanh(S \cdot \kappa_{ij})$
28:        $w'_{ij} \leftarrow w_{ij} \cdot (1 - \text{suppress} \cdot 0.9)$  ▷ **Suppress Redundancy**
29:    **end if**
30: **end for**
   // *Phase 4: Optimization*
31: Update graph $G$ with new weights $W'$.
32: $Y \leftarrow$ OptimizeCrossEntropy($Y$, $G$)
33: **return** $Y$

# 4 Experiments

This section evaluates the performance of the proposed algorithm through a series of experiments. We first analyze two typical failure modes of standard UMAP on representative manifold datasets, namely topological tearing and structural collapse. We then compare the proposed method with mainstream dimensionality reduction approaches across multiple datasets. Finally, through hyperparameter studies, we verify the specific role of the Jaccard mechanism within the algorithm.

## 4.1 Analysis of Topological Tearing and Structural Collapse

We select two classical three-dimensional manifolds, the Trefoil Knot and the Swiss Roll, as test datasets to analyze the phenomena of topological tearing and structural collapse. Specifically, the Trefoil Knot is used to examine structural collapse, while the Swiss Roll is employed to investigate topological tearing.

### 4.1.1 Trefoil Knot: Structural Collapse

Figure 1 illustrates the original distribution of the Trefoil Knot data in three-dimensional space, together with the corresponding two-dimensional embeddings produced by standard UMAP and JORC-UMAP. The Trefoil Knot is the simplest non-trivial knot, and its local neighborhoods contain pronounced regions of high curvature.

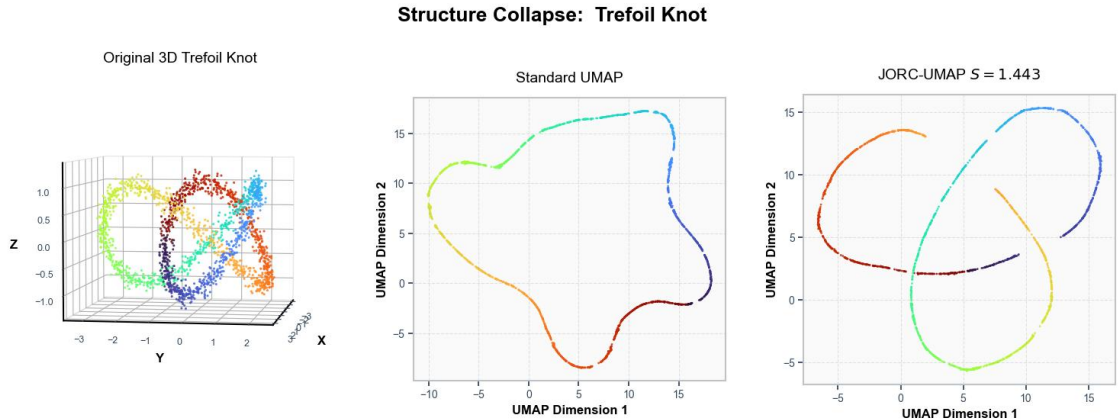

Figure 1: Structural collapse experiment on the Trefoil Knot dataset

From the embedding produced by standard UMAP, it can be observed that although global connectivity is largely preserved, distinct branches that are well separated in the original three-dimensional space undergo noticeable local compression and even overlap in the two-dimensional representation. In particular, regions of high curvature  where separation should be maintained —are drawn unnaturally close to one another, resulting in an unreasonable contraction of the overall structure. We refer to this phenomenon as *structural collapse*.

In contrast, the embedding generated by JORC-UMAP on the same dataset preserves the overall shape of the Trefoil Knot and the relative relationships among its branches to a much greater extent, with no evident collapse occurring in high-curvature regions. This comparison indicates that, on manifolds with pronounced curvature variation, neighborhood graphs constructed solely from local Euclidean distances are insufficient, and that explicitly incorporating local geometric information at the graph construction stage helps alleviate the resulting structural distortions.

### 4.1.2 Swiss Roll: Topological Tearing Phenomenon

Figure 2 presents the embeddings of the Swiss Roll dataset obtained by different methods. The Swiss Roll is a canonical example of a two-dimensional manifold embedded in three-dimensional space, whose intrinsic structure requires a dimensionality reduction method to preserve local neighborhood relationships while simultaneously respecting the global unfolding order of the manifold.

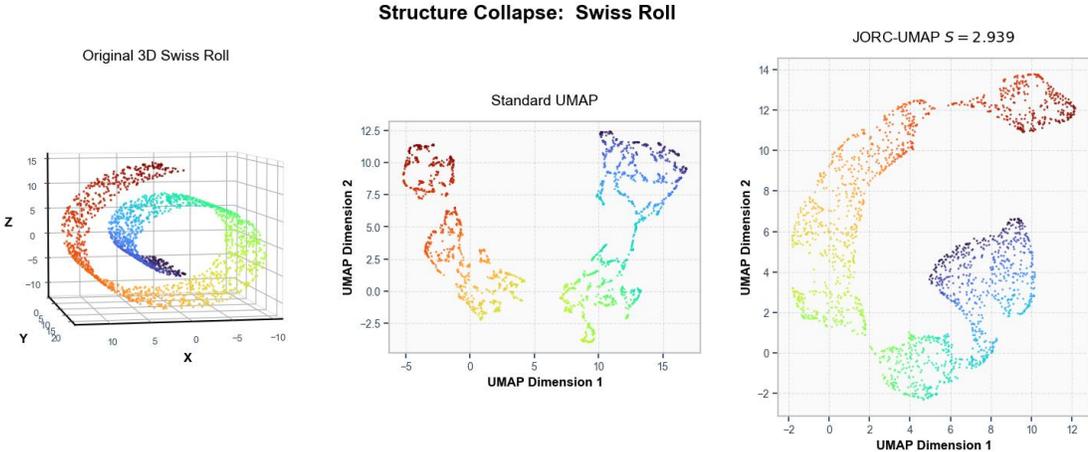

Figure 2: Swiss Roll experiment for evaluating topological tearing

In the embedding produced by standard UMAP, the originally continuous manifold is incorrectly fragmented into multiple disconnected components. Although the local structure within each fragment is largely preserved, the absence of meaningful connections between these components prevents the global continuity of the manifold from being maintained in the low-dimensional space. This behavior exemplifies the topological tearing phenomenon, where an intrinsically connected manifold is erroneously severed during the embedding process.

In contrast, the embedding generated by JORC-UMAP on the Swiss Roll dataset largely restores the global continuity of the manifold. The relative ordering and connectivity between different regions are preserved more consistently, and no pronounced disconnections are observed. This result further indicates that, when manifolds exhibit long-range dependencies or non-uniform unfolding structures, neighborhood graphs constructed solely from local distance information are prone to disrupting global connectivity, whereas incorporating curvature-aware structural cues helps mitigate such topological distortions.

Based on the experimental results on both the Trefoil Knot and Swiss Roll datasets, it can be observed that standard UMAP exhibits two characteristic failure modes on manifolds with pronounced curvature or complex topology, namely structural collapse and topological tearing, respectively. These deficiencies are unlikely to stem primarily from the low-dimensional optimization stage; instead, they are more plausibly attributed to limitations in the neighborhood graph construction process, where the geometric and connectivity properties of the data manifold are insufficiently captured. This observation further underscores the necessity of explicitly incorporating geometric and topological information at the graph construction stage. Subsequent experiments will therefore systematically evaluate the extent to which JORC-UMAP mitigates these issues under more general data conditions.

## 4.2 Numerical Experiments

### 4.2.1 Experimental Design

We conduct a comparative evaluation of JORC-UMAP against t-SNE, UMAP, TriMap, and PaCMAP. It has been well established that dimensionality reduction (DR) methods exhibit pronounced sensitivity to hyperparameter choices, and that no single parameter configuration can simultaneously achieve optimal performance across diverse datasets [27, 5]. Accordingly, for each algorithm we consider three representative hyperparameter settings. For each evaluation metric, we report the best result attained by each method across these settings; moreover, to accommodate variations in dataset scale and distribution, hyperparameters are allowed to undergo limited, dataset-specific adjustments, strictly confined to the recommended ranges. The concrete hyperparameter configurations are summarized as follows:

1. **t-SNE**: perplexity$\in$ 10, 20, 40；

2. **UMAP**: number of neighbors($n_{NB}$)$\in$ 10, 15, 20；

3. **TriMap**: number of inliers($n_{inlier}$)$\in$ 8, 10, 15；

4. **PaCMAP**: number of neighbors($n_{NB}$)$\in$ 5, 10, 20；

5. **JORC-UMAP**: Jaccard threshold $\delta \in 0.05, 0.1, 0.15$, and number of neighbors($n_{NB}$)$\in$ 10, 15, 20。

The test datasets are organized into four groups according to their experimental objectives and intrinsic characteristics:

1. MNIST[12] and Fashion-MNIST[28]. These datasets feature well-defined class labels and relatively smooth manifolds with clear class structures. They are primarily used to assess an algorithm's ability to separate classes and preserve local neighborhood relationships on standard visual data.

2. 20 Newsgroups [11], Olivetti Faces [21], and the United States Postal Service (USPS) handwritten digits dataset [8]. This group is employed to examine the preservation of semantic clustering structures, as well as the trade-off between maintaining local neighborhood fidelity and capturing global organization.

3. S-curve, Trefoil Knot, and the Mammoth dataset [25].These datasets are designed to analyze the structural preservation capabilities of dimensionality reduction methods under conditions of high curvature or nontrivial topological complexity.

4. Noisy S-curve and noisy Swiss Roll datasets.These datasets are used to evaluate the robustness of the algorithms to noise, specifically their ability to recover the true manifold backbone while suppressing spurious noise-induced connections.

### 4.2.2 Algorithm Evaluation

Following the evaluation framework and metric selection principles proposed in PaCMAP [26], this study adopts three complementary quantitative criteria to assess the performance of different dimensionality reduction methods.Here we detail the three evaluation metrics used in our experiments:

1. **SVM Accuracy** A support vector machine (SVM) with an RBF kernel [22] is employed, and the classification performance is evaluated using 5-fold cross-validation. This metric is mainly used to measure the compactness of class-wise neighborhoods in the low-dimensional embedding. Its evaluation does not directly depend on local density estimation in the embedding space.

2. **Random Triplet Accuracy** Randomly sampled Triplets $(i, j, k)$ , and the proportion of triplets whose relative distance ordering is consistent between the high-dimensional space and the low-dimensional embedding is computed. This metric is used to evaluate the preservation of overall distance relationships. To control computational cost, this procedure is repeated five times and the average value is reported.

3. **Centroid Triplet Accuracy** centroids are computed in both the high-dimensional and low-dimensional spaces, and triplets are constructed based on the relative distances between these centroids. This metric is used to assess whether the relative positions of different classes as a whole are preserved after dimensionality reduction.

Among these metrics, SVM Accuracy mainly reflects the preservation of local structure in the low-dimensional embedding, while Random Triplet Accuracy and Centroid Triplet Accuracy are used to evaluate the preservation of global structure.

### 4.2.3 Experimental Results

An important advantage of dimensionality reduction and visualization methods is that their effectiveness can be directly examined through the visual inspection of low-dimensional embeddings [6].

The two-dimensional embedding results for the MNIST and Fashion-MNIST datasets are shown in Figure 3. The results for the 20 Newsgroups, Olivetti Faces, and USPS handwritten digit datasets are shown in Figure 4.

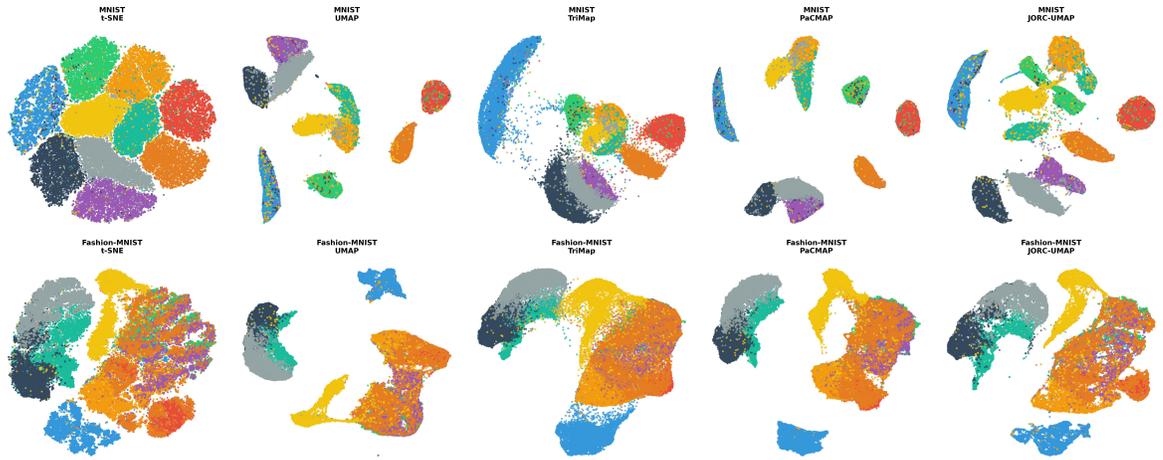

Figure 3: Two-dimensional embedding results on the MNIST and Fashion-MNIST datasets

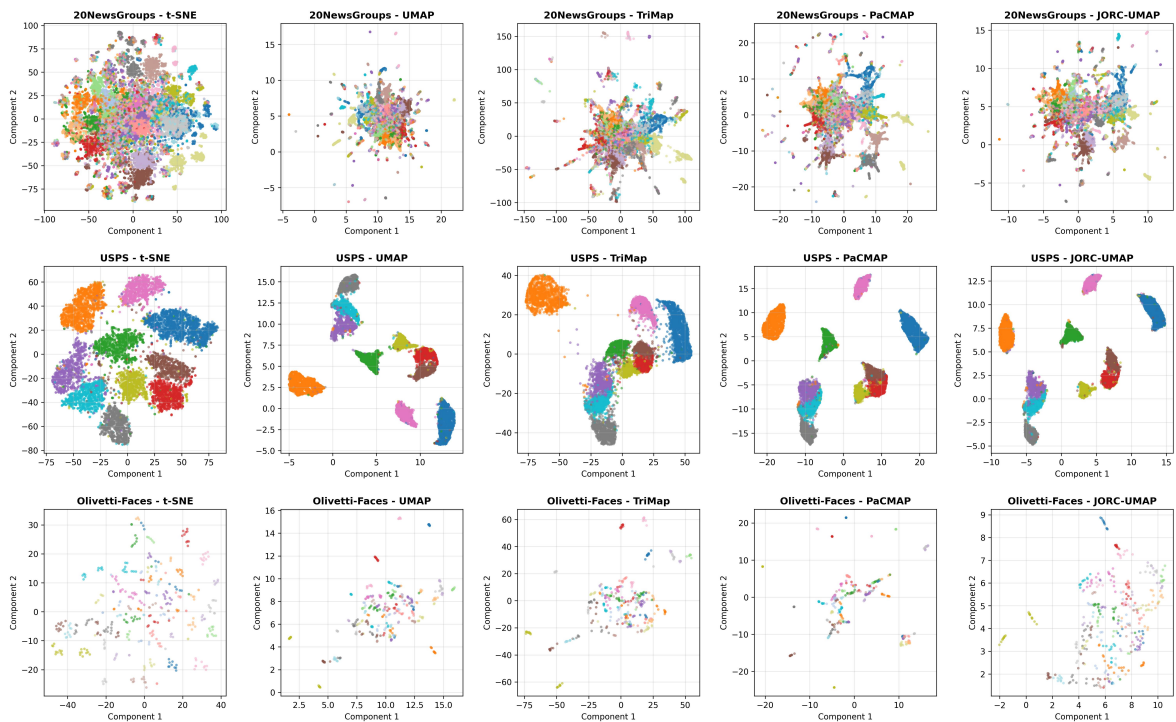

Figure 4: Two-dimensional embedding results on the 20 Newsgroups, Olivetti Faces, and USPS handwritten digit datasets

Figures 3-4 compare the two-dimensional embeddings produced by different dimensionality reduction methods on image, text, and handwritten digit datasets. t-SNE tends to spread samples more uniformly across the embedding space, which weakens global inter-class relationships. Standard UMAP places greater emphasis on preserving local neighborhood structure. TriMap and PaCMAP strike a compromise between global and local structure preservation, but their embeddings still exhibit a certain degree of distortion. In contrast, JORC-UMAP produces an overall structure similar to that of PaCMAP, while further enlarging the separation between different classes, resulting in clearer inter-class relationships.

The experimental results on the S-curve, Trefoil Knot, and Mammoth datasets are shown in Figure 5.Figure 6 presents the two-dimensional embedding results of different dimensionality reduction methods on the noisy S-curve and noisy Swiss Roll datasets.

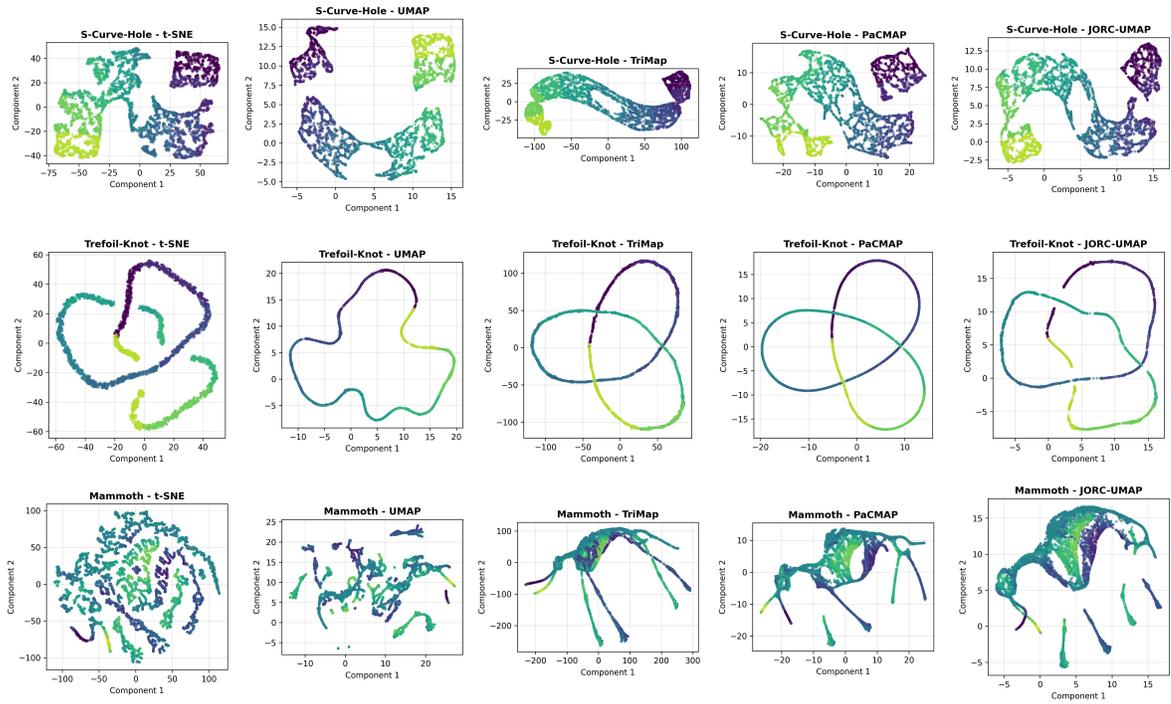

Figure 5: Two-dimensional embedding results on the S-curve, Trefoil Knot, and Mammoth datasets.

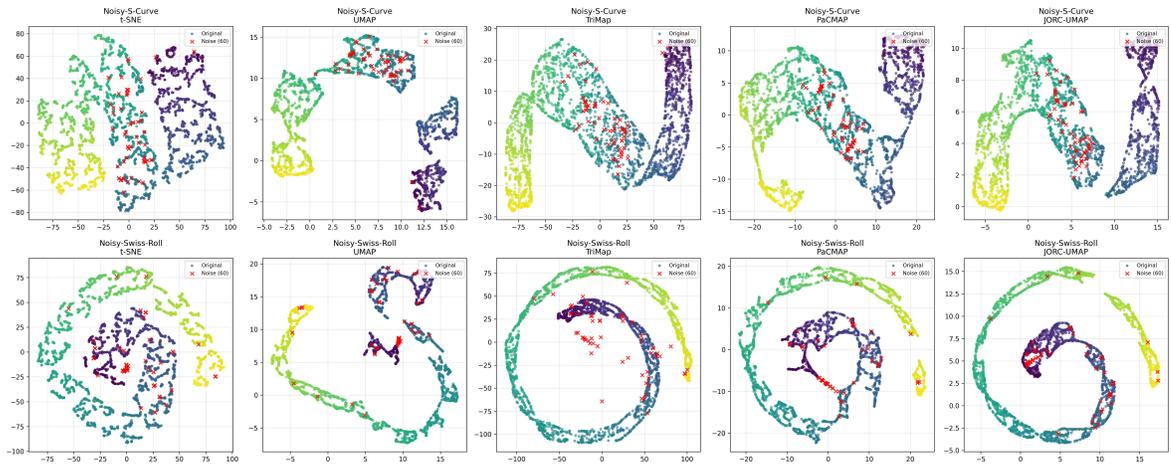

Figure 6: Two-dimensional embedding results on noisy datasets.

Figure 5 compares the two-dimensional embeddings produced by different dimensionality reduction methods on the S-curve, Trefoil Knot, and Mammoth datasets. In the noise-free setting, t-SNE and standard UMAP exhibit fragmentation on the S-curve and structural collapse in high-curvature regions of the Trefoil Knot. In contrast, TriMap, PaCMAP, and JORC-UMAP better preserve the overall shape of the manifolds, with JORC-UMAP retaining both local details and the global outline on the Mammoth dataset. Figure 6 reports results on noisy datasets, where t-SNE and standard UMAP are more susceptible to spurious neighborhood relationships, and PaCMAP tends to preserve non-structural connections introduced by noise. By comparison, TriMap and JORC-UMAP effectively suppress such connections while maintaining the continuity of the manifold backbone. Overall, the incorporation of ORC and the Jaccard similarity improves structural stability under complex geometry and noisy conditions.

Table 1 reports SVM Accuracy, Table 2 presents Random Triplet Accuracy, and Table 3 shows Centroid Triplet Accuracy.

Table 1: SVM Accuracy on Different Datasets

| **Dataset** | **t-SNE** | **UMAP** | **TriMap** | **PaCMAP** | **JORC-UMAP** |
|---|---|---|---|---|---|
| MNIST | 0.965 | 0.967 | 0.960 | **0.970** | 0.960 |
| Fashion-MNIST | 0.743 | 0.742 | 0.730 | **0.748** | 0.746 |
| 20NewsGroups | 0.447 | 0.422 | 0.425 | **0.464** | 0.442 |
| Olivetti Faces | **0.648** | 0.573 | 0.562 | 0.589 | 0.555 |
| USPS | 0.951 | **0.956** | 0.941 | 0.950 | 0.949 |

Table 2: Random Triplet Accuracy (RTE) on Different Datasets

| **Dataset** | **t-SNE** | **UMAP** | **TriMap** | **PaCMAP** | **JORC-UMAP** |
|---|---|---|---|---|---|
| MNIST | **0.625** | 0.615 | 0.608 | 0.616 | 0.620 |
| Fashion-MNIST | 0.747 | 0.736 | **0.776** | 0.741 | 0.774 |
| 20NewsGroups | 0.640 | 0.659 | **0.695** | 0.654 | 0.635 |
| Olivetti Faces | **0.719** | 0.708 | 0.710 | 0.655 | 0.706 |
| USPS | 0.664 | 0.660 | 0.644 | **0.665** | 0.662 |

Table 3: Centroid Triplet Accuracy (CTE) on Different Datasets

| **Dataset** | **t-SNE** | **UMAP** | **TriMap** | **PaCMAP** | **JORC-UMAP** |
|---|---|---|---|---|---|
| MNIST | 0.695 | 0.716 | 0.673 | **0.749** | 0.724 |
| Fashion-MNIST | 0.781 | 0.810 | **0.854** | 0.810 | 0.841 |
| 20NewsGroups | 0.780 | 0.777 | **0.808** | 0.764 | 0.792 |
| Olivetti Faces | **0.765** | 0.751 | 0.737 | 0.683 | 0.741 |
| USPS | 0.772 | 0.797 | 0.746 | 0.803 | **0.814** |

In terms of SVM Accuracy (Table 1), PaCMAP and t-SNE achieve higher scores on most supervised datasets, indicating a stronger emphasis on local class separability. The performance of JORC-UMAP is generally close to that of the best-performing methods, without explicitly enforcing extreme intra-class compression.For Random Triplet Accuracy (Table 2), TriMap outperforms other methods on most datasets, reflecting its strong constraints on global distance ordering. JORC-UMAP attains results close to the best method on multiple datasets, demonstrating stable global structure preservation.Regarding Centroid Triplet Accuracy (Table 3), JORC-UMAP achieves the best or second-best performance on datasets such as USPS, suggesting an advantage in preserving class-level geometric relationships while not significantly sacrificing local structure.

Overall, across the three metrics, JORC-UMAP does not aim to optimize a single criterion but instead exhibits a balanced trade-off between local discriminability and global structural consistency. This behavior aligns with its design goal of improving the overall stability of the embedding through curvature awareness and neighborhood consistency constraints, rather than optimizing for a specific evaluation metric.

## 4.3 Subsection: Validation of the Necessity of the Jaccard Coefficient and Hyperparameter Sensitivity Analysis

To systematically evaluate the role of the Jaccard similarity coefficient and the resulting core control parameter—the Jaccard threshold —in JORC-UMAP, this section designs a controlled synthetic experiment to assess how this parameter affects the embedding results.

### 4.3.1 Validation of the Necessity of the Jaccard Coefficient

To examine the effect of the Jaccard similarity coefficient in the proposed algorithm, we construct a synthetic dataset consisting of three concentric rings connected by narrow noisy bridges. This data structure contains both genuine manifold connections and noise-induced shortcut connections, making it suitable for testing the algorithm's ability to distinguish between them. The experimental results are shown in Figure 7. In all experiments, the strength parameter S is fixed to 2.

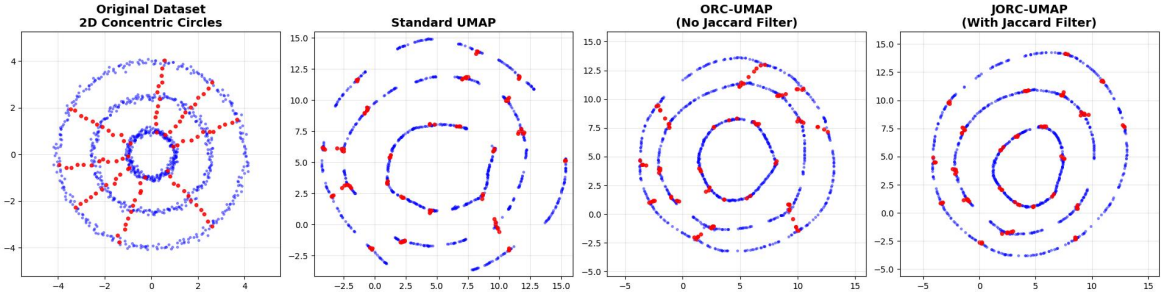

Figure 7: Jaccard-based noise robustness results on the three-ring bridging dataset. The Original Dataset shows the true geometric structure in the high-dimensional space. Standard UMAP denotes the method without curvature or Jaccard-based adjustment. ORC-UMAP enhances edge weights solely based on Ollivier-Ricci curvature. JORC-UMAP further incorporates Jaccard similarity for topological filtering on top of ORC-UMAP, with the threshold set to δ = 0.1.

As shown in Figure 7, standard UMAP exhibits clear topological tearing between the three rings and preserves several non-genuine bridging paths induced by noise. ORC-UMAP partially restores connectivity between the rings by strengthening negative-curvature edges; however, it may also amplify noise-induced connections, leading to more pronounced noisy bridges than those observed with standard UMAP. In contrast, after introducing the Jaccard similarity coefficient, JORC-UMAP effectively suppresses noise edges caused by low neighborhood overlap and selectively enhances only those connections with highly shared neighborhoods. While preserving global topological connectivity, it substantially reduces interference from noise channels and more robustly mitigates topological tearing. These results indicate that when curvature information alone may amplify noise-induced connections, incorporating a Jaccard-based neighborhood overlap constraint helps distinguish structural negative curvature from noise-induced negative curvature, thereby improving the stability of the resulting embeddings.

### 4.3.2 Hyperparameter Sensitivity Analysis of the Jaccard Threshold δ

This subsection further examines the effect of the Jaccard threshold δ on the embedding results. By gradually increasing δ on the 3D S-curve dataset, we analyze how this parameter controls the trade-off between structural preservation and noise suppression. The experimental results are shown in Figure 8, where the strength parameter S is fixed to 2.

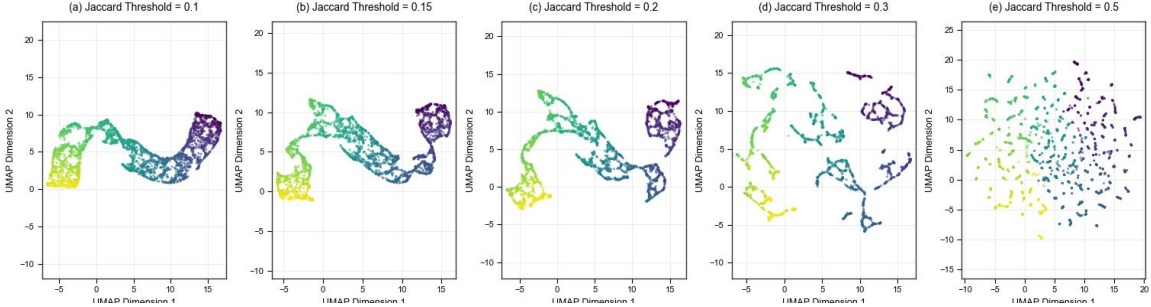

Figure 8: Hyperparameter study of the Jaccard similarity threshold , with  set to 0.1, 0.15, 0.2, 0.3, and 0.5.

Figure 8 illustrates the effect of the Jaccard similarity threshold δ on the embedding results produced by JORC-UMAP. When δ = 0.1, the threshold is relatively low, and most negative-curvature edges pass the Jaccard test, leading to a relatively permissive selection of potential backbone connections. In this case, the embedding exhibits good overall continuity but still retains some weak connections induced by noise.

As δ increases to 0.15 and 0.2, the algorithm reaches a more favorable balance between noise suppression and backbone preservation. On the one hand, noise-induced bridging edges with low neighborhood overlap are effectively weakened; on the other hand, key negative-curvature edges induced by the manifold geometry are preserved, resulting in embeddings that are continuous, smooth, and consistent with the original S-curve topology.

When δ is further increased to 0.3 and 0.5, the Jaccard test becomes overly restrictive, and the algorithm begins to globally suppress negative-curvature edges. Since true manifold bottleneck regions often lie in low-density areas with limited neighborhood overlap, some structural connections may be misclassified as noise and weakened, leading to clear fragmentation and discontinuities in the embedding, as well as a noticeable degradation of global structure preservation.

Overall, δ acts as a discriminative scale in JORC-UMAP for distinguishing structural negative-curvature edges from noise-induced negative-curvature edges. Within a reasonable intermediate range, the algorithm exhibits good robustness with respect to δ, and the observed performance gains primarily arise from the joint effect of curvature information and Jaccard-based constraints, rather than from fine-grained tuning of a specific hyperparameter.

# 5 Discussion and Conclusion

The proposed JORC-UMAP primarily targets two common failure modes of standard UMAP on complicated manifold data—erroneous disconnection of topological tearing and structural collapse and provides a unified improvement scheme. By incorporating Ollivier-Ricci curvature (ORC) into the neighborhood graph construction stage to modulate edge weights, the method preserves local structure while more effectively maintaining the global manifold structure, thereby alleviating these failure phenomena across a variety of datasets.

The core methodological innovation lies in a rethinking of how curvature is used. Owing to the fundamental differences in dimensionality and geometric structure between the high-dimensional data space and the low-dimensional embedding space, directly aligning or comparing their curvatures within the optimization objective lacks a sound theoretical basis. JORC-UMAP therefore does not attempt to impose explicit curvature constraints in the embedding space. Instead, it introduces Ollivier-Ricci curvature into the discrete k-nearest neighbor graph to modulate edge weights. This strategy avoids the issue of curvature incomparability across spaces while strengthening structurally important connections prior to embedding. Since methods such as PaCMAP and TriMap also rely on neighborhood graph construction, the proposed approach can, in principle, serve as a general-purpose module that integrates naturally with a broad range of existing manifold learning and visualization algorithms based on k-nearest neighbor graph.

Considering that ORC may amplify connections induced by noise, this work further introduces the Jaccard similarity coefficient to filter negative curvature edges. This mechanism exploits local neighborhood overlap information to effectively distinguish key negative curvature edges arising from genuine geometric structure from spurious connections caused by noise, thereby enhancing global structure preservation while improving the stability of the algorithm in noisy settings.

Experimental results demonstrate that JORC-UMAP achieves stable and consistent improvements across a wide range of scenarios, including synthetic manifolds, image data, text data, and single-cell datasets. These findings provide empirical evidence supporting the effectiveness of the proposed method.

It should be noted that computing ORC introduces additional computational overhead, which becomes particularly pronounced in large-scale data scenarios. An important future direction is to first use lightweight criteria to quickly identify key edges that are likely to affect the global structure, and then selectively compute curvature only on these edges to reduce the overall computational cost. In addition, different types of discrete curvature have their own advantages in characterizing data structure. Designing more efficient and structure-adaptive discrete curvature computation and approximation methods tailored to different data characteristics would further broaden the application of curvature-based ideas in dimensionality reduction and visualization algorithms.

# Acknowledgments

Xiaobin Li is supported by NSFC grant No. 11501470, No. 11426187, No. 11791240561, the Fundamental Research Funds for the Central Universities 2682021ZTPY043 and partially supported by NSFC grant No. 11671328. Especially, Xiaobin Li would like to thank Bohui Chen, An-min Li, Guosong Zhao for their constant support and also thank all the friends met in different conferences.